\pdfoutput=1

\documentclass[11pt]{article}

\usepackage{acl}

\usepackage[utf8]{inputenc} 
\usepackage[T1]{fontenc}    
\usepackage{hyperref}       
\usepackage{url}            
\usepackage{booktabs,colortbl}       
\usepackage{amsfonts}       
\usepackage{nicefrac}       
\usepackage{microtype}      
\usepackage{xcolor}         

\usepackage{graphics}
\usepackage{graphicx}
\usepackage{subfig}
\usepackage{enumitem}
\usepackage{balance}

\usepackage{times}
\usepackage{latexsym}

\usepackage{times}
\usepackage{latexsym}
\usepackage{bm}
\usepackage{amssymb}
\usepackage{booktabs,colortbl}       
\usepackage{multirow}
\usepackage{arydshln}

\usepackage{ascii}

\definecolor{zred}{RGB}{196, 38, 11}
\definecolor{zblue}{RGB}{41, 52, 190}
\definecolor{zgreen}{RGB}{18, 141, 21}

\definecolor{zptu}{RGB}{18, 141, 21}

\definecolor{wxjiao}{RGB}{18, 21, 141}

\title{Understanding and Improving Sequence-to-Sequence Pretraining for Neural Machine Translation}

\author{
Wenxuan Wang$^1$\thanks{\ \ Work was mainly done when Wenxuan Wang and Yongchang Hao were interning at Tencent AI Lab.}  \quad Wenxiang Jiao$^{2}$ \quad Yongchang Hao$^{3}$ \quad Xing Wang$^{2}$ \\ 
 \bf Shuming Shi$^2$ \quad Zhaopeng Tu$^2$\thanks{\ \ Zhaopeng Tu is the corresponding author.} \quad Michael R. Lyu$^1$\\
$^1$Department of Computer Science and Engineering, The Chinese University of Hong Kong \\ {\asciifamily \normalsize \{wxwang,lyu\}@cse.cuhk.edu.hk} \\
$^2$Tencent AI Lab \\  {\asciifamily \normalsize \{joelwxjiao,brightxwang,shumingshi,zptu\}@tencent.com} \\
$^3$University of Alberta \qquad  {\asciifamily \normalsize yongcha1@ualberta.ca} \\
}

\begin{document}
\maketitle
\begin{abstract}

In this paper, we present a substantial step in better understanding the SOTA sequence-to-sequence (Seq2Seq) pretraining for neural machine translation~(NMT). We focus on studying the impact of the jointly pretrained decoder, which is the main difference between Seq2Seq pretraining and previous encoder-based pretraining approaches for NMT. 
By carefully designing experiments on three language pairs, we find that Seq2Seq pretraining is a double-edged sword: On one hand, it helps NMT models to produce more diverse translations and reduce adequacy-related translation errors. On the other hand, the discrepancies between Seq2Seq pretraining and NMT finetuning limit the translation quality (i.e., domain discrepancy) and induce the over-estimation issue (i.e., objective discrepancy).
Based on these observations, we further propose simple and effective strategies, named in-domain pretraining and input adaptation to remedy the domain and objective discrepancies, respectively. Experimental results on several language pairs show that our approach can consistently improve both translation performance and model robustness upon Seq2Seq pretraining. 
\end{abstract}

\section{Introduction}

There has been a wealth of research over the past several years on self-supervised pre-training for natural language processing tasks~\cite{Devlin2019BERT,liu2019roberta,Conneau2020xlmr,jiao2020exploiting}, which aims at transferring the knowledge of large-scale unlabeled data to downstream tasks with labeled data.
Despite its success in other understanding and generation tasks, self-supervised pretraining is not a common practice in machine translation (MT).
One possible reason is the architecture discrepancy between pretraining model (e.g., Transformer {\bf encoder}) and NMT models (e.g., Transformer {\bf encoder-decoder}). 

To remedy the architecture gap, several researchers propose sequence-to-sequence (Seq2Seq) pretraining models for machine translation, e.g., MASS~\cite{song2019mass} and BART~\cite{zhu2019incorporating,lewis2020bart}.
Recently,~\newcite{Liu2020mbart} extend BART by training on large-scale multilingual language data (i.e., mBART), leading to significant improvement on translation performance across various language pairs.
While previous pretraining approaches for NMT generally focus only on Transformer encoder~\cite{Lample2019xlm}, mBART pretrains a complete autoregressive Seq2Seq model by recovering the input sentences that are noised by masking phrases.
One research question naturally arises: {\em how much does the jointly pretrained decoder matter?}

In this work, we present a substantial step in better understanding the SOTA Seq2Seq pretraining model. We take a fine-grained look at the impact of the jointly pretrained decoder by carefully designing experiments, which are conducted on several WMT and IWSLT benchmarks across language pairs and data scales using the released mBART-25 model~\cite{Liu2020mbart}. 
By carefully examining the translation outputs, we find that ($\S~\ref{sec:translation-performance}$):
\begin{itemize}[leftmargin=10pt]
    \item Jointly pretraining decoder produces more diverse translations with different word orders, which calls for multiple references to accurately evaluate its effectiveness on large-scale data.
    \item Jointly pretraining decoder consistently reduces adequacy-related translation errors over pretraining encoder only.
\end{itemize}

Although jointly pretraining decoder consistently improves translation performance, we also identify several side effects due to the discrepancies between pretraining and finetuning (\S\ref{sec:pretrain-finetune-gaps}):
\begin{itemize}[leftmargin=10pt]
    \item {\bf domain discrepancy}: Seq2Seq pretraining model is generally trained on general domain data while the downstream translation models are trained on specific domains (e.g., news). The domain discrepancy requires more efforts for the finetuned model to adapt the knowledge in pretrained models to the target in-domain.
    \item {\bf objective discrepancy}: NMT training learns to translate a sentence from one language to another, while Seq2Seq pretraining learns to reconstruct the input sentence. The objective discrepancy induces the over-estimation issue and tends to generate more hallucinations with noisy input. The over-estimation problem along with more copying translations induced by Seq2Seq pretraining~\cite{Liu:2021:ACL} make it suffer from more serious beam search degradation problem. 
\end{itemize}

To remedy the above discrepancies, we propose simple and effective strategies, named  in-domain pretraining and input adaptation in finetuning (\S\ref{sec:approach}). In in-domain pretraining, we propose to reduce the domain shift by continuing the pretraining of mBART on in-domain monolingual data, which is more similar in data distribution with the downstream translation tasks.
For input adaptation, we add noises to the source sentence of bilingual data, and combine the noisy data with the clean bilingual data for finetuning. We expect the perturbed inputs to better transfer the knowledge from pretrained model to the finetuned model.
Experimental results on the benchmark datasets show that in-domain pretraining improves the translation performance significantly and input adaptation enhances the robustness of NMT models. Combining the two approaches gives us the final solution to a well-performing NMT system. Extensive analyses show that our approach can narrow the domain discrepancy, particularly improving the translation of low-frequency words. Besides, our approach can alleviate the over-estimation issue and mitigate the beam search degradation problem of NMT models.

\begin{table*}[t!]
\centering
\setlength{\tabcolsep}{5pt}
\begin{tabular}{ccc cc cc cc cc cc cc}
\toprule
\multicolumn{3}{c}{\bf Pretraining}  
& \multicolumn{2}{c}{\bf W19 En-De}  
& \multicolumn{2}{c}{\small \bf W19 En-De (S)}  
& \multicolumn{2}{c}{\bf W16 En-Ro}
& \multicolumn{2}{c}{\bf I17 En-Fr} \\
\cmidrule(lr){1-3} \cmidrule(lr){4-5}  \cmidrule(lr){6-11} 
\em Model   &   \em {Enc} & \em {Dec} 
&  \em $\Rightarrow$  &   \em $\Leftarrow$ &  \em $\Rightarrow$  &   \em $\Leftarrow$ &  \em $\Rightarrow$  &   \em $\Leftarrow$ &  \em $\Rightarrow$  &   \em $\Leftarrow$  \\
\midrule
\multicolumn{3}{c}{\bf no pretrain}  &  39.6    &   41.0   & 29.7 & 30.1   & 34.5 &   34.3   & 37.3 & 38.0   \\
\midrule
\multirow{3}{*}{\bf mBART}  &   \texttimes  &   \texttimes     & 39.4  & 40.1  & 26.7 &  27.1  &   30.0 & 29.6 & 35.3 & 35.1 \\
   & \checkmark    &   \texttimes  & \cellcolor{red!11.2} 40.8 & \cellcolor{red!8} 41.1  &  \cellcolor{red!64} 31.7 & \cellcolor{red!51.2} 33.5 & \cellcolor{red!40} {35.0} & \cellcolor{red!48} {35.6} & 
  \cellcolor{red!62} 38.4 & \cellcolor{red!66}38.4 \\
  &   \checkmark    &   \checkmark   & \cellcolor{red!11.2} 40.8 & \cellcolor{red!10.4} 41.4 & {\cellcolor{red!}} 35.3 & {\cellcolor{red!84.6}} 35.7 &  \cellcolor{red!60} 37.1 & \cellcolor{red!62.4} 37.4 &
\cellcolor{red!80}39.2 & \cellcolor{red!99}40.2 \\
\bottomrule
\end{tabular}
\caption{BLEU scores on MT benchmarks. ``Enc:\texttimes, Dec:\texttimes'' represents that we use only the pre-trained embeddings for fair comparisons, and we highlight performance improvement over this setting in {\color{zred}red} color.}
\label{tab:benchmarks}
\end{table*}

\section{Understanding Seq2Seq Pretraining}

In this section, we conduct experiments and analyses to gain a better understanding of current Seq2Seq pretraining for NMT. We first present the translation performance of the pretrained components (\S\ref{sec:translation-performance}), and then show the discrepancy between pretraining and finetuning (\S\ref{sec:pretrain-finetune-gaps}). 

\subsection{Experimental Setup}

\paragraph{Data.} 
We conduct experiments on several benchmarks across language pairs, including high-resource WMT19 English-German (W19 En-De, 36.8M instances), and low-resource WMT16 English-Romanian (W16 En-Ro, 610K instances) and IWSLT17 English-French (I17 En-Fr, 250K instances). 
To eliminate the effect of different languages, we also sample a subset from WMT19~En-De (i.e., W19~En-De~(S), 610K instances) to construct a low-resource setting for ablation studies.

For the proposed {\em in-domain pretraining}, we collect the NewsCrawl monolingual data as the in-domain data for WMT tasks (i.e., 200M English, 200M German, and 60M Romanian), and the TED monolingual data for IWSLT tasks (i.e., 1M English and 0.9M French). Since the monolingual data from TED is rare, we expand it with pseudo in-domain data, OpenSubtitle~\cite{tiedemann2016finding}, which also provides spoken languages as TED. Specifically, we use the latest 200M English subtitles and all the available French subtitles (i.e., 100M).
We follow~\newcite{Liu2020mbart} to use their released sentence-piece model~\cite{kudo-richardson-2018-sentencepiece} with 250K subwords to tokenize both bilingual and monolingual data.
We evaluate the translation performance using the SacreBLEU~\cite{post-2018-call}.

\paragraph{Models.}
As for the pretrained models, we adopt the officially released mBART25 model~\cite{Liu2020mbart}\footnote{\url{https://github.com/pytorch/fairseq/tree/main/examples/mbart}}, which is trained on the large-scale CommonCrawl~(CC) monolingual data in 25 languages.
As a result, the vocabulary is very large in mBART25, including 250K words.
mBART uses a larger Transformer model which extends both the encoder and decoder of Transformer-Big to 12 layers. We use the parameters of either encoder or encoder-decoder from the pretrained mBART25 for finetuning. Then, in the following section, we use pretrained encoder, and pretrained encoder-decoder for short.
We follow the officially recommended finetuning setting with dropout of $0.3$, label smoothing of $0.2$, and warm-up of $2500$ steps. We finetune on the high-resource task for 100K steps and the low-resource tasks for 40K steps, respectively.

We also list the results of vanilla Transformer without pretraining as baseline.
The vocabulary is built on the bilingual data, hence is much smaller (e.g., En-De 44K) than mBART25.
Specifically, for high-resource tasks we train 6L-6L Transformer-Big with 460K tokens per batch for 30K steps, and for low-resource tasks we train 6L-6L Transformer-Base with 16K tokens per batch for 50K steps.

\subsection{Impact of Jointly Pretrained Decoder}
\label{sec:translation-performance}

The main difference of Seq2Seq pretraining models (e.g., mBART) from previous pretraining models (e.g., BERT and XLM-R) lies in whether to train the decoder together.
In this section, we investigate the impact of the jointly pretrained decoder in terms of BLEU scores, and provide some insights on where the jointly pretrained decoder improves performance.

\paragraph{Translation Performance.}
Table~\ref{tab:benchmarks} lists the BLEU scores of pretraining different components of NMT models, where we also include the results of NMT models trained on the datasets from scratch (``no pretrain''). 
For fair comparisons, we use the same vocabulary size for all variants of pretraining NMT components.
We use the pretrained word embedding for the model variant with randomly initialized encoder-decoder (``Enc:\texttimes, Dec:\texttimes''), which makes it possible to train 12L-12L NMT models on the small-scale datasets. Accordingly, the results of (``Enc:\texttimes, Dec:\texttimes'') is worse than the ``no pretrain'' model due to the larger vocabulary (e.g., 250K vs. 44K) that makes the model training more difficult. 

Pretraining encoder only (``Enc:\checkmark, Dec:\texttimes'') significantly improves translation performance, which is consistent with the findings in previous studies~\cite{zhu2019incorporating,weng2020acquiring}.
{We also conduct experiments with the pretrained encoder XLM-R~\cite{Conneau2020xlmr}, which achieves comparable performance as the mBART encoder (see Appendix~\ref{sec:XLMR}). For fair comparisons, we only use the mBART encoder in the following sections.}
Encouragingly, jointly pretraining decoder can further improve translation performance, although the improvement is not significant on the large-scale WMT19 En-De data. These results seem to provide empirical support for the common cognition -- pretraining is less effective on large-scale data. However, we have some interesting findings of the generated outputs, which may draw different conclusions. 
To eliminate the effect of language and data bias, we use the full set and sampled subset of WMT19 De$\Rightarrow$En data as representative large-scale and small-scale data scenarios.

\begin{table}[t]
\centering
\small
\setlength{\tabcolsep}{5pt}
\begin{tabular}{c m{6cm}}
\toprule
\em Src  & Sie bezichtigt die Erwachsenen Kinderhandel zu betreiben.  \\
\em Ref   &  She accuses the adults of child trafficking. \\
\midrule
\multicolumn{2}{c}{\bf Large-Scale Data}\\
no pre. &  {\color{red}It} accuses {\color{red}(the)} adults of children trafficking. \\
(\texttimes, \texttimes)    &  {\color{red}It} accuses {\color{red}(the)} adults of children trafficking.  \\
(\checkmark, \texttimes)    &  She accuses the adults of children trafficking. \\
(\checkmark, \checkmark)    & She accuses the adults of {\color{blue}trafficking in children}.
  \\
\midrule
\multicolumn{2}{c}{\bf Small-Scale Data}\\
no pre. &  {\color{red}It} accuses the adults {\color{red}to trade} children. \\
(\texttimes, \texttimes) &  {\color{red}It requires adult trafficking on children. }\\
(\checkmark, \texttimes) & {\color{red}It} accuses {\color{red}(the)} adults of children trafficking. \\
(\checkmark, \checkmark)    &  She accuses the adults of {\color{blue}trafficking in children}. \\
\bottomrule
\end{tabular}
\caption{Translation examples on WMT19 De$\Rightarrow$En test set. The translation errors are highlighted in {\color{red}red} and changes of word order are highlighted in {\color{blue}blue}.}
\label{tab:case-study}
\end{table}

Table~\ref{tab:case-study} shows some translation examples. Firstly, jointly pretraining decoder can produce good translations that are different in the word order from the ground-truth reference (e.g., ``trafficking in children'' vs. ``child trafficking"), thus are assigned low BLEU scores. 
This may explain why jointly pretraining decoder only marginally improves performance on large-scale data.
Secondly, jointly pretraining decoder can reduce translation errors, especially on small-scale data (e.g., correct the mistaken translation of ``{\em It}'' to ``{\em She}'').
We empirically validate the above two findings in the following experiments.

\begin{table}[t]
\centering
\fontsize{10}{12}\selectfont
\begin{tabular}{c lr lr}
\toprule
\multirow{2}{*}{\bf Pretrain}   &   \multicolumn{2}{c}{\bf Single}   &   \multicolumn{2}{c}{\bf Multiple}\\
\cmidrule(lr){2-3}  \cmidrule(lr){4-5}
  &   BLEU  &  $\triangle$ &   BLEU    &   $\triangle$ \\
\midrule
\multicolumn{5}{l}{\bf Large-Scale Data}\\
no pretrain               &   39.5    & -   &   77.1    &   -\\
\hdashline
(\texttimes, \texttimes)  &   38.6    & -0.9    &   75.7    &   -1.4\\
(\checkmark, \texttimes)  &   39.5    & +0.0    &   77.8    &   +0.7\\
(\checkmark, \checkmark)  &   39.9    & +0.4    &   79.1$^{\Uparrow}$    &   \bf +2.0\\
\midrule
\multicolumn{5}{l}{\bf Small-Scale Data}\\
no pretrain   &  27.0    &  -   &   53.1    &   -\\
\hdashline
(\texttimes, \texttimes)  &  27.0   &   +0.0 &  52.3    &   -0.8\\
(\checkmark, \texttimes)  &  32.3   &   +5.3 &  63.4    &   +10.3\\
(\checkmark, \checkmark)  &  35.3$^{\Uparrow}$    &   +8.3    &   69.1$^{\Uparrow}$    &   \bf +16.0\\
\bottomrule
\end{tabular}
\caption{BLEU scores on En$\Rightarrow$De testset with single and multiple references. ``$\Uparrow$'' denotes significantly better (with $p < 0.01$) than No mBART pretraining.}
\label{tab:multi-reference}
\end{table}

\paragraph{Impact on Translation Diversity.}
We follow~\newcite{Du:2021:ICML} to better evaluate the translation quality for different word orders using multiple references. 
We use the test set released by~\citet{ott2018analyzing}, which consists of 10 human translations for 500 sentences taken from the WMT14 En$\Rightarrow$De test set.
As shown in Table~\ref{tab:multi-reference}, the pretrained decoder achieves more significant improvement in all cases when measured by multiple references. These results provide empirical support for our claim that jointly pretraining decoder produces more diverse translations with different word orders, which can be better measured by multiple references.
These results may renew our cognition of pretraining, that is, {\em they are also effective on large-scale data when evaluated more accurately}.

\begin{table}[t!]
\fontsize{10}{12}\selectfont
\centering
\begin{tabular}{cc rr rr rr}
\toprule
\multicolumn{2}{c}{\bf Pretrain}   &   \multicolumn{3}{c}{\bf Large}   &   \multicolumn{3}{c}{\bf Small}\\
\cmidrule(lr){1-2}  \cmidrule(lr){3-5}  \cmidrule(lr){6-8}
\em {Enc} & \em {Dec} & \em Ut   &   \em Mt &  \em Ot & \em Ut   &   \em Mt &  \em Ot\\
\cmidrule(lr){1-2}  \cmidrule(lr){3-8}
\texttimes  &   \texttimes & 4 & 9 & 0 & 25 & 45 & 0 \\
\checkmark  &   \texttimes & 3 & 3 & 0 & 5 & 21 & 5 \\
\checkmark  &   \checkmark & 2 & 0 & 0 & 3 & 15 & 0 \\
\bottomrule
\end{tabular}
\caption{Human evaluation of mBART pretrained NMT models in terms of under-translation (Ut), mis-translation (Mt), and over-translation (Ot) errors.}
\label{tab:mbart-adequacy}
\end{table}

\paragraph{Impact on Adequacy.}
We conduct a human evaluation to provide a more intuitive understanding of how jointly pre-training decoder improves translation quality. Specifically, we ask two annotators to annotate under-translation, mis-translation and over-translation on 100 sentences randomly sampled from WMT19 De$\Rightarrow$En test set. As listed in Table~\ref{tab:mbart-adequacy}, inheriting the pretrained decoder reduces more translation errors on small data than on large data, which is consistent with the results of BLEU score in Table~\ref{tab:benchmarks}. 
Interestingly, inheriting only the pretrained encoder introduces more over-translation errors on small data, which can be solved by combining the pretrained decoder. One possible reason is that inheriting only the pretrained encoder excessively enlarges the impact of source context.\footnote{\newcite{Tu:2017:TACL} showed that more impact of source context leads to over-translation errors.} 
This problem does not happen on large data, since the large amount of in-domain data can balance the relation between encoder and decoder to accomplish the translation task well.

\subsection{Pretraining-and-Finetuning Discrepancy}
\label{sec:pretrain-finetune-gaps}

Although Seq2Seq pretraining consistently improves translation performance across data scales, we find several side effects of Seq2Seq pretraining due to the discrepancy between pretraining and finetuning.
In this section, we present two important discrepancies: {\em domain discrepancy} and {\em objective discrepancy}. Unless otherwise stated, we report results on WMT19 En-De test set using small data.

\subsubsection{Domain Discrepancy}
\label{sec:domain-discrepancy}

Seq2Seq pretraining model is generally trained on {\bf general domain} data while the downstream translation models are trained on {\bf specific domains} (e.g., news). Such a domain discrepancy requires more efforts for the finetuned models to adapt the knowledge in pretrained models to the target in-domain. We empirically show the domain discrepancy in terms of lexical distribution and domain classifier.

\begin{figure}[t!]
    \centering
    \includegraphics[height=0.3\textwidth]{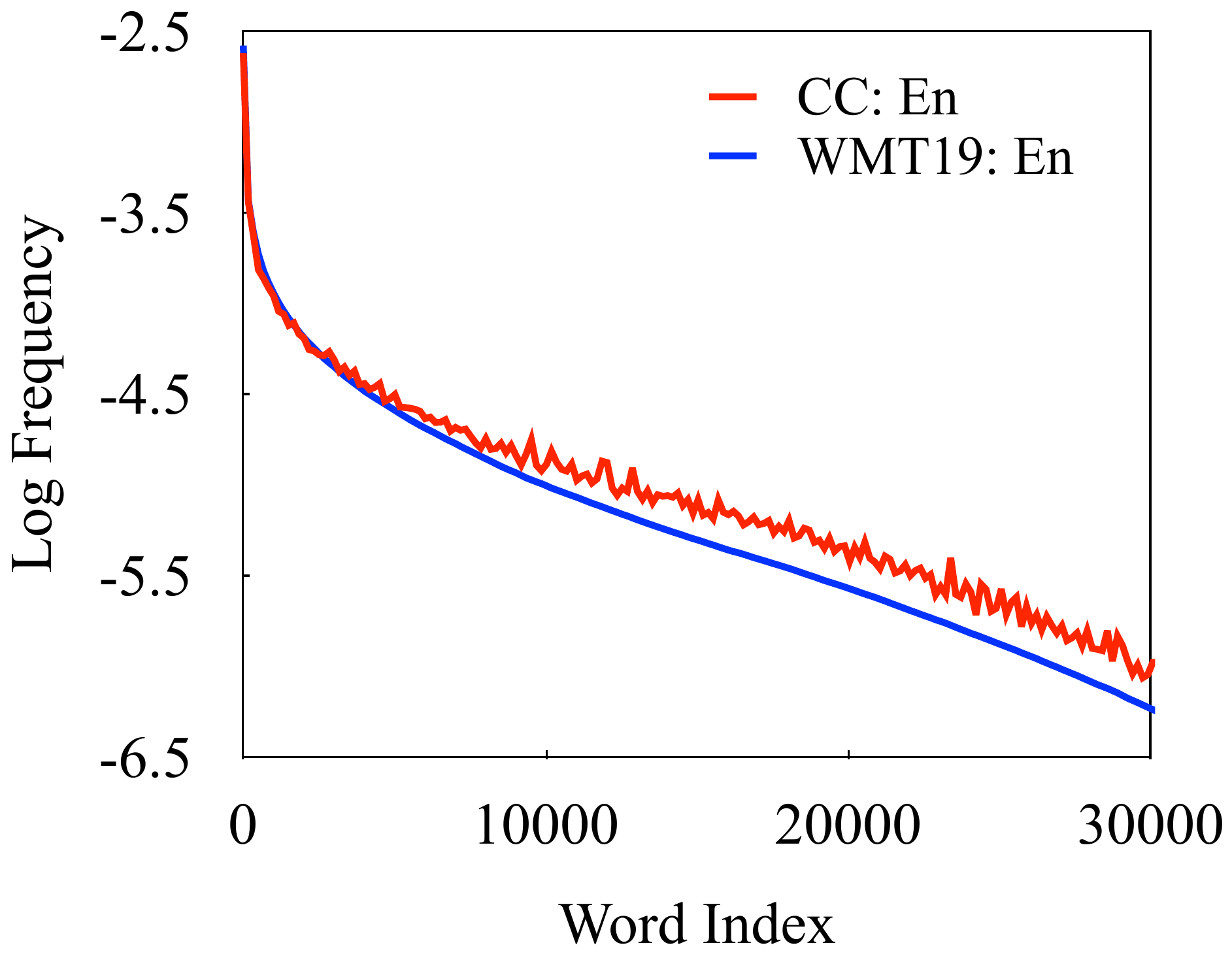}
    \caption{Word distributions of English corpora from general domain (i.e., CC data) and in-domain (i.e., WMT19 En-De news domain), respectively. The word frequency is normalized and reported in log-scale.}
    \label{fig:distribution}
\end{figure}

\paragraph{Lexical Distribution in Training Data.}
Inspired by lexicon distribution analysis~\cite{ding2021understanding}, we first plot the word distributions of English corpora from general domain (i.e., CC data) and in-domain (i.e., WMT19 En-De news domain) to study their difference at the lexicon level. The words are ranked according to their frequencies in the WMT19 En-De training data. As shown in Figure~\ref{fig:distribution}, we observe a clear difference between WMT news data and CC data in the long tail region, which is supposed to carry more domain-specific information. Accordingly, there will be a domain shift from pretraining to finetuning.

\begin{table}[t!]
\fontsize{10}{12}\selectfont
  \centering
  \begin{tabular}{c cc cc}
  \toprule
  \bf Set
  & \bf En$\Rightarrow$De 
  & \bf De$\Rightarrow$En \\
  \midrule
   Source & 77.5 & 73.7 \\
   Target & 71.0 &  75.4\\
  \bottomrule
  \end{tabular}
  \caption{Ratio of sentences in WMT19 En-De test sets that are classified as WMT news domain.}
  \label{tab:domain-cls}
\end{table}

\paragraph{Domain Classifier for Test Data.}
We further demonstrate that the test data also follows a consistent domain as the training data. To distinguish general domain and in-domain, we build a domain classifier based on the WMT19 En-De training data and the CC data. 
We select a subset from the WMT training data with some trusted data~\cite{wang2018denoising,jiao2020data,jiao2022data2}, which includes 22404 sample from WMT newstest2010-2017 {(see Appendix~\ref{sec:domain-classifier} for details).}
Specifically, we select 1.0M samples from the WMT training data and the CC data, respectively, to train the domain classifier.
The newstest2018 is combined with an equally sized subset of CC data for validation. We adopt the domain classifier to classify each sample in the test sets of WMT19 En-De. As shown in Table~\ref{tab:domain-cls}, most of the sentences (e.g., 70\% - 80\%) are recognized as WMT news domain, which demonstrates the domain consistency between the training data and test data in the downstream tasks.

\subsubsection{Objective Discrepancy}
\label{sec:obj-discrepancy}

The learning objective discrepancy between Seq2Seq pretraining and NMT training is that NMT learns to translate a sentence from one language to another, while Seq2Seq pretraining learns to reconstruct the input sentence~\cite{Liu:2021:ACL}. 
In this section, we study the side effects of the objective discrepancy by evaluating the predicting behaviors that are highly affected by the learning objective.

\begin{figure}[t!]
    \centering  
    \subfloat[Reference]{
    \includegraphics[height=0.26\textwidth]{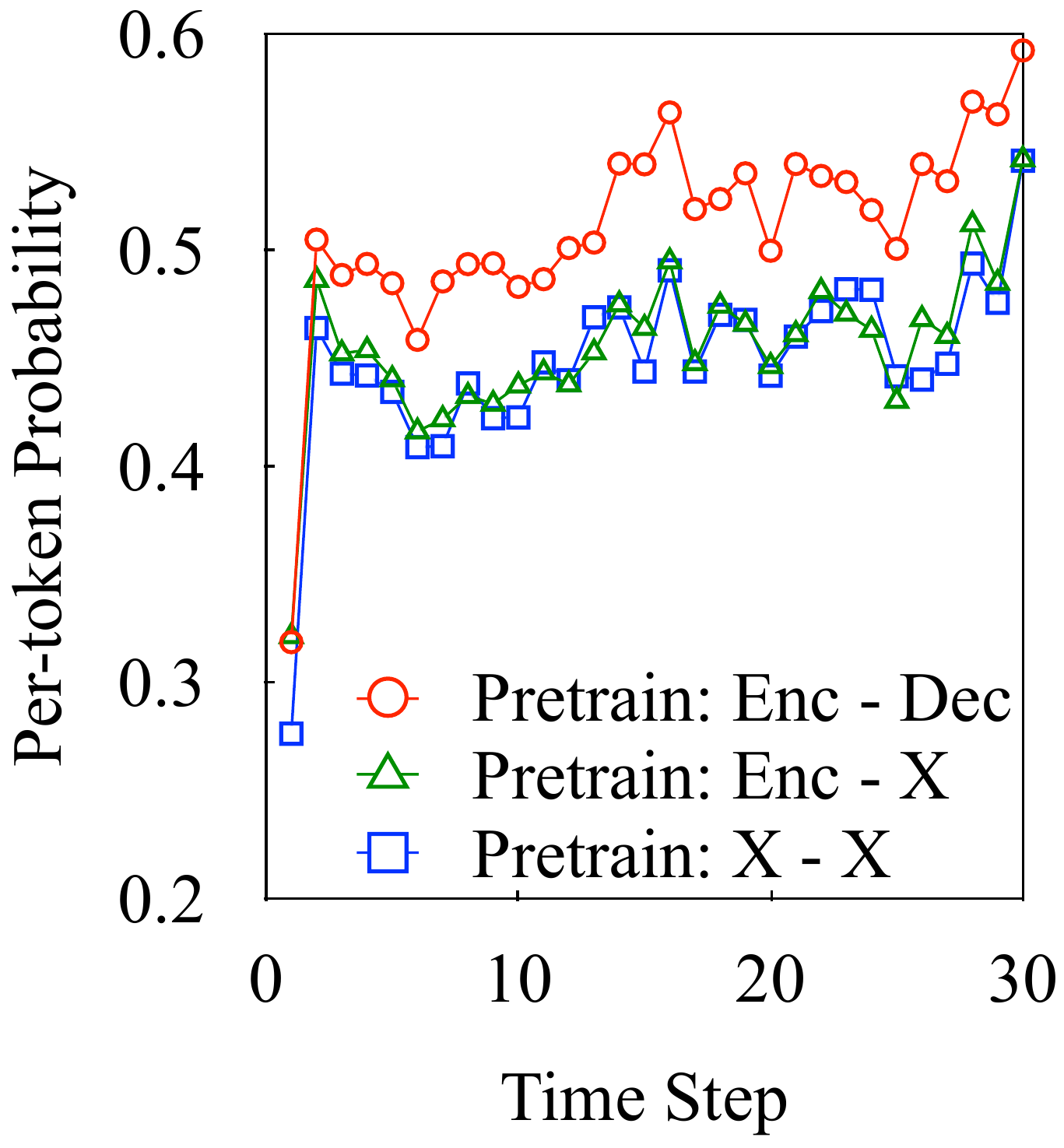}}
    \hfill
    \subfloat[Distractor]{
    \includegraphics[height=0.26\textwidth]{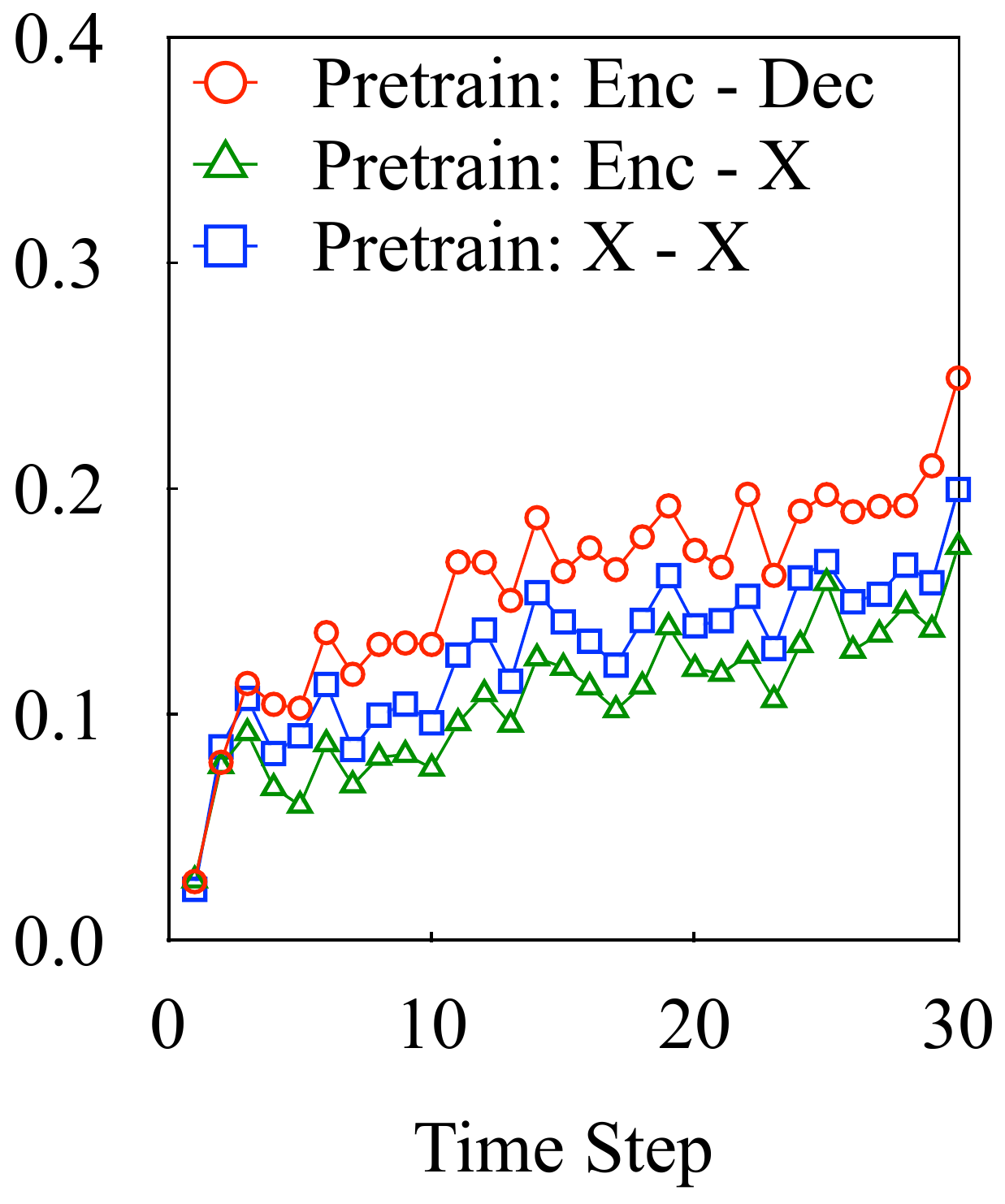}}
    \caption{Per-token generation probability on the test set of WMT19 En$\Rightarrow$De~(S). Higher probabilities are expected for the groundtruth references (a), and lower probabilities are expected for the distractors (b).}
    \label{fig:model-uncert}
\end{figure}

\paragraph{Model Uncertainty.}
We follow~\newcite{ott2018analyzing} to analyze the model's uncertainty by computing the average probability at each time step across a set of sentence pairs. 
To evaluate the capability of LM modeling on the target language, we also follow~\newcite{wang2020exposure} to consider a set of ``distractor'' translations, which are random sentences from the CC data that match the corresponding reference translation in length.
Figure~\ref{fig:model-uncert} plots model uncertainties for both references ($Y$) and distractors ($\hat{Y}$).
We find that jointly pretraining decoder significantly improves model certainty after the first few time steps (Figure~\ref{fig:model-uncert}a).
As for the distractors, pretraining encoder only results in certainties even lower than training from scratch (Figure~\ref{fig:model-uncert}b), which suggests that the corresponding NMT model is more dominated by the source context. It reconfirms the finding in our human evaluation (Table~\ref{tab:mbart-adequacy}). In contrast, jointly pretraining decoder leads to a significant improvement of certainties, suggesting that the pretrained decoder tends to induce the {\bf over-estimation issue} of NMT models. A possible reason is that Seq2Seq pretraining does not establish the connection between languages, such that its strong capability of LM modeling still recognizes the distractor as a valid target sentence even though it is mismatched with the source sentence in semantics.

\paragraph{Hallucination under Perturbation.}
One translation problem associated with over-estimation is hallucination~\cite{wang2020exposure}, where NMT models generate fluent translation but is unrelated to the input. In this section, we follow~\newcite{lee2018hallucinations} to evaluate the model's tendency of generating hallucination under noisy input, to which NMT models are highly sensitive~\cite{belinkov2018synthetic}.
Specifically, we employ two different perturbation strategies: (1) First position insertion (FPI) that inserts a single additional input token into the source sequence, which can completely divorce the translation from the input sentence~\cite{lee2018hallucinations}.
(2) Random span masking (RSM) that simulates the noisy input in the Seq2Seq pretraining of mBART~\cite{Liu2020mbart}.
We follow ~\newcite{lee2018hallucinations} to count a translation as hallucination under perturbation (\textsc{HuP}) when: (1) BLEU between reference sentence and translation of unperturbed sentence is bigger than 5 and (2) BLEU between the translation of perturbed sentence and the translation of unperturbed sentence is lower than 3. We calculate the percentage of hallucination as the \textsc{HuP} score. Table~\ref{tab:mbart-halluc} lists the BLEU change and \textsc{HuP} score for the perturbed inputs. As expected, jointly pretraining decoder is less robust to perturbed inputs (more decline of BLEU scores), and produces more hallucinations than the other two model variants.

\begin{table}[t!]
\fontsize{10}{12}\selectfont
\centering
\begin{tabular}{cc rr rr}
\toprule
\multicolumn{2}{c}{\bf Pretrain}   
& \multicolumn{2}{c}{\bf FPI (\%)}
& \multicolumn{2}{c}{\bf RSM (\%)}\\
\cmidrule(lr){1-2}  \cmidrule(lr){3-4} \cmidrule(lr){5-6}
\em {Enc} & \em {Dec} 
& \em  $\triangle_{\mathrm{BLEU}}$ & \em \textsc{HuP}
& \em $\triangle_{\mathrm{BLEU}}$ & \em \textsc{HuP} \\
\midrule
\texttimes  &   \texttimes & -1.3 & 0.5 & -8.8 & 2.4  \\
\checkmark  &   \texttimes & -0.3 & 0.5 & -8.3 & 0.5  \\
\checkmark  &   \checkmark & \bf -3.2 & \bf 7.8 & \bf -17.8 & \bf 15.5 \\
\bottomrule
\end{tabular}
\caption{BLEU change of model performance under perturbed inputs over the standard inputs, and hallucinations under perturbation (HUP) score.}
\label{tab:mbart-halluc}
\end{table}

\begin{table}[t!]
\fontsize{10}{12}\selectfont
\centering
\begin{tabular}{cc cc cc}
\toprule
\multicolumn{2}{c}{\bf Pretrain}   
& \multicolumn{2}{c}{\bf BLEU}
& \multicolumn{2}{c}{\bf Copy (\%)}\\
\cmidrule(lr){1-2}  \cmidrule(lr){3-4} \cmidrule(lr){5-6}
\em {Enc} & \em {Dec} 
& 5 & 100 & 5 & 100\\
\midrule
\texttimes  &   \texttimes & 26.7 & \cellcolor{zgreen!5}26.6 & 12.9 & 13.9 \\
\checkmark  &   \texttimes & 31.7 & \cellcolor{zgreen!5}31.6 & 12.7 & 12.9 \\
\checkmark  &   \checkmark & 35.3 & \cellcolor{zgreen!90}33.5 & \bf 13.2 & \bf 19.4 \\
\bottomrule
\end{tabular}
\caption{Beam search degradation and ratio of copying tokens in translation outputs.}
\label{tab:mbart-BSD}
\end{table}

\paragraph{Beam Search Problem.} 
One commonly-cited weakness of NMT model is the beam search problem, where the model performance declines as beam size increases~\cite{Tu:2017:AAAI}.
Previous studies demonstrate that over-estimation is an important reason for the beam search problem~\cite{ott2018analyzing,cohen2019empirical}.
We revisit this problem for NMT models with Seq2Seq pretraining, as shown in Table~\ref{tab:mbart-BSD}.
We also list the ratio of copying tokens in translation outputs (i.e., directly copy source words to target side without translation) for different beam sizes, which has been shown as a side effect of Seq2Seq pretraining models~\cite{Liu:2021:ACL}.
As seen, jointly pretraining decoder suffers from more serious beam search degradation problem, which reconfirms the connection between beam search problem and over-estimation.
In addition, larger beam size introduces more copying tokens than the other model variants (i.e., 19.4 vs. 13.9, 12.9), which also links copying behaviors associated with Seq2Seq pretraining to the beam search problem.

\section{Improving Seq2Seq Pretraining}
\label{sec:approach}

\subsection{Approach}

To bridge the above gaps between Seq2Seq pretraining and finetuning, we introduce {\em in-domain pretraining} and {\em input adaptation} to improve the translation quality and model robustness.

\paragraph{In-Domain Pretraining.}
To bridge the domain gap, we propose to continue the training of mBART~\cite{Liu2020mbart} on the in-domain monolingual data. Specifically, we first remove
spans of text and replace them with a mask token. We mask 35\% of the words in each sentence by random sampling a span length according to a Poisson distribution ($\lambda=3.5$). We also permute
the order of sentences within each instance. The training objective is to reconstruct the original sentence at the target side. 
We expect the in-domain pretraining to reduce the domain shift by re-pretraining on the in-domain data, which is more similar in data distribution with the downstream translation tasks. 

\paragraph{Input Adaptation in Finetuning.}
To bridge the objective gap and improve the robustness of models, we propose to add noises (e.g., mask, delete, permute) to the source sentences during finetuning, and keep target sentences as original ones. Empirically, we add noises to 10\% of the words in each source sentence, and combine the noisy data with the clean data by the ratio of 1:9, which are used to finetune the pretraining model.
We expect the introduction of perturbed inputs in finetuning can help to better transfer the knowledge from pretrained model to the finetuned model, thus alleviate over-estimation and improve the model robustness.

\begin{table*}[t!]
  \centering
  \begin{tabular}{l cr cr cr cr}
  \toprule
  \multirow{2}{*}{\bf Approach} 
  & \multicolumn{2}{c}{\bf W19 En$\Rightarrow$De} 
  & \multicolumn{2}{c}{\bf \small W19 En$\Rightarrow$De (S)} 
  & \multicolumn{2}{c}{\bf W16 En$\Rightarrow$Ro} 
  & \multicolumn{2}{c}{\bf I17 En$\Rightarrow$Fr} \\
  \cmidrule(lr){2-3} \cmidrule(lr){4-9}
 & BLEU & \textsc{HuP} &  BLEU & \textsc{HuP} & BLEU & \textsc{HuP} &  BLEU & \textsc{HuP} \\
  \midrule
  Baseline   & 39.4 & 2.6  & 26.7 & 2.4  & 30.0 &  1.1 & 35.3 & 1.6 \\
  General   & 40.8 & 3.3  & 35.3 & 15.5  & 37.1 &  6.5 & 39.2 & 7.8 \\
  \hline 
 ~~~+ Input Adapt & 40.8 & 2.7 &  35.6 &  5.7 & 37.2 & 2.4 & 39.4 & 1.5 \\
  ~~~+ In-Domain & \bf 42.2 & 9.2   & \bf 36.4 &  10.4  & \bf 38.0 &  8.2  & 39.9 &  5.5  \\
  ~~~~~~+ Input Adapt & 41.3  & 4.1 &  36.1  & 3.6 & 37.8 & 2.9 & \bf 40.1 & 3.0\\
  \bottomrule
  %
  \multirow{2}{*}{\bf Approach} 
  & \multicolumn{2}{c}{\bf W19 De$\Rightarrow$En} 
  & \multicolumn{2}{c}{\bf \small W19 De$\Rightarrow$En (S)} 
  & \multicolumn{2}{c}{\bf W16 Ro$\Rightarrow$En} 
  & \multicolumn{2}{c}{\bf I17 Fr$\Rightarrow$En} \\
  \cmidrule(lr){2-3} \cmidrule(lr){4-9}
 & BLEU & \textsc{HuP} &  BLEU & \textsc{HuP} & BLEU & \textsc{HuP} &  BLEU & \textsc{HuP} \\
  \midrule
  Baseline &  40.1  & 2.8  &27.1   &  1.3 &  29.6 & 1.3   & 35.1 & 1.7 \\
  General   & \bf 41.4 & 7.7 & 35.7 &  4.9  &  37.4 &  6.0 &  40.2 &  4.7 \\
  \hline
  ~~~+ Input Adapt & 41.2 & 2.6 &  35.9 &  2.8 & 37.1 & 3.5 & 40.7 & 2.5 \\
  ~~~+ In-Domain & 41.3  &  8.2  & \bf 36.9 &    7.4  & \bf 38.1 & 7.7 & \bf 41.1  & 4.2\\
  ~~~~~~+ Input Adapt & \bf 41.4  &3.1 &  36.8  & 2.9 & 37.9 & 3.9 & 41.0 & 1.7\\
  \bottomrule
  \end{tabular}
  \caption{
  BLEU and \textsc{HuP} scores of our approaches for downstream translation tasks.}
  \label{tab:main-results}
\end{table*}

\subsection{Experimental Results}

\paragraph{Main Results on Translation Performance and Robustness.}
The main results are listed in Table~\ref{tab:main-results}. We report the results of input adaptation, in-domain pretraining, and the combination of these two approaches, respectively. For input adaptation, it achieves comparable translation quality as the general domain pretrained model and significantly reduces the ratio of \textsc{HuP}, indicating the enhancement of model robustness. 
In-domain pretraining generally improves the translation quality but does not make the model more robust. On the contrary, it may increase the ratio of \textsc{HuP} in some cases (e.g., En$\Rightarrow$Ro 5.6 vs. 8.2). Conducting input adaptation right after in-domain pretraining will combine the advantages of these two approaches, and improve both the translation quality and model robustness. 
{The effectiveness of our approaches, especially input adaptation, is more significant when evaluated with multiple references, as shown in Table~\ref{tab:multi-ref-impr}. }

\begin{table}[t!]
\setlength{\tabcolsep}{4pt}
\fontsize{10}{12}\selectfont
\centering
  \begin{tabular}{l cc cc}
  \toprule
  \bf Approach
  & \multicolumn{2}{c}{\bf W19 En-De}
  & \multicolumn{2}{c}{\bf \small W19 En-De~(S)} \\
  \cmidrule(lr){2-3}  \cmidrule(lr){4-5}
&   BLEU  &  $\triangle$ &   BLEU    &   $\triangle$  \\
  \midrule
   Baseline & 75.7  & - &  52.3 & - \\
   General & 79.1 & +3.4  &  69.1 & +16.8 \\
   ~~~+ Input Adapt &  79.2 & +3.5 &  71.7 & +19.4\\
   ~~~+ In-Domain & \bf 80.1 & \bf +4.4 &  73.7 & +21.4\\
   ~~~~~~+ Input Adapt & 79.8 &  +4.1 & \bf 75.6 & \bf +23.3\\
  \bottomrule
  \end{tabular}
  \caption{BLEU scores with multiple references.}
  \label{tab:multi-ref-impr}
\end{table}

\paragraph{In-Domain Only.}
Given the promising performance of in-domain pretraining, we investigate whether pretraining on in-domain data only can also obtain significant improvement.
We report the results in Table~\ref{tab:indomain-only}. We can observe that pretraining solely on the in-domain data can improve the translation performance noticeably over the models without pretraining. However, the improvement is less competitive than the pretrained mBART25 (e.g., En$\Rightarrow$Ro: 36.1 v.s. 37.1 in Table~\ref{tab:main-results}), which may result from the much larger scale of multilingual data used in general pretraining.

\begin{table}[t!]
\setlength{\tabcolsep}{4pt}
\fontsize{10}{12}\selectfont
  \centering
  \begin{tabular}{l cc cc}
  \toprule
  \bf Approach
  & \multicolumn{2}{c}{\bf\small W19 En-De (S)}
  & \multicolumn{2}{c}{\bf W16 En-Ro}\\
  \cmidrule(lr){2-3} \cmidrule(lr){4-5}
  &  $\Rightarrow$  &   $\Leftarrow$  &  $\Rightarrow$  & $\Leftarrow$ \\
  \midrule
   Baseline & 26.7 & 27.1 & 30.0 & 29.6 \\
   In-Domain & 35.2 & 35.7 & 36.1 & 36.3 \\
  \bottomrule
  \end{tabular}
  \caption{BLEU scores of in-domain pretraining only.}
  \label{tab:indomain-only}
\end{table}

\subsection{Analysis}
We provide some insights into how our approach improves model performance over general pretraining. We report results on WMT19 En$\Rightarrow$De test set using small-scale data.

\paragraph{Narrowing Domain Gap.}
Since the difference of lexical distribution between general domain and in-domain data mainly lies in the long tail region (see Figure~\ref{fig:distribution}), we study how our approach performs on low-frequency words.
Specifically, we calculate the word accuracy of the translation outputs for WMT19 En-De~(S) by the \texttt{compare-mt}\footnote{\url{https://github.com/neulab/compare-mt}} tool. We follow previous studies~\cite{Wang:2021:ACL,Jiao2021SelfTrainingSW} to divide words into three categories based on their frequency in the bilingual data, including High: the most {3,000} frequent words; Medium: the most {3,001}-{12,000} frequent words; Low: the other words.
Table~\ref{tab:linguistic-freq} lists the results. 
The improvements on low-frequency words are the major reason for the performance gains of in-domain pretraining, where it outperforms general pretraining on the translation accuracy of low/medium/high- frequency words by 1.7, 0.0, and 0.7 BLEU scores, respectively.
These findings confirm our hypothesis that in-domain pretraining can narrow the domain gap with in-domain data, which is more similar in the lexical distribution as the test sets.

\begin{table}[t!]
\fontsize{10}{12}\selectfont
  \centering
  \begin{tabular}{l ccc}
  \toprule
  \multirow{2}{*}{\bf Approach}  &  \multicolumn{3}{c}{\bf Frequency} \\
  \cmidrule(lr){2-4}
   & \it Low  & \it Med &  \it High \\
  \midrule
   Baseline  & 36.8 & 45.3 & 57.5  \\
   General  & 44.5 & 54.3 & 64.2 \\
   ~~+ In-Domain & 46.2 & 54.3 & 64.9 \\
  \bottomrule
  \end{tabular}
  \caption{F-measures of word prediction for different frequencies that are calculated in the bilingual data.}
  \label{tab:linguistic-freq}
\end{table}

\begin{figure}[t]
    \centering  
    \subfloat[Reference]{
    \includegraphics[height=0.26\textwidth]{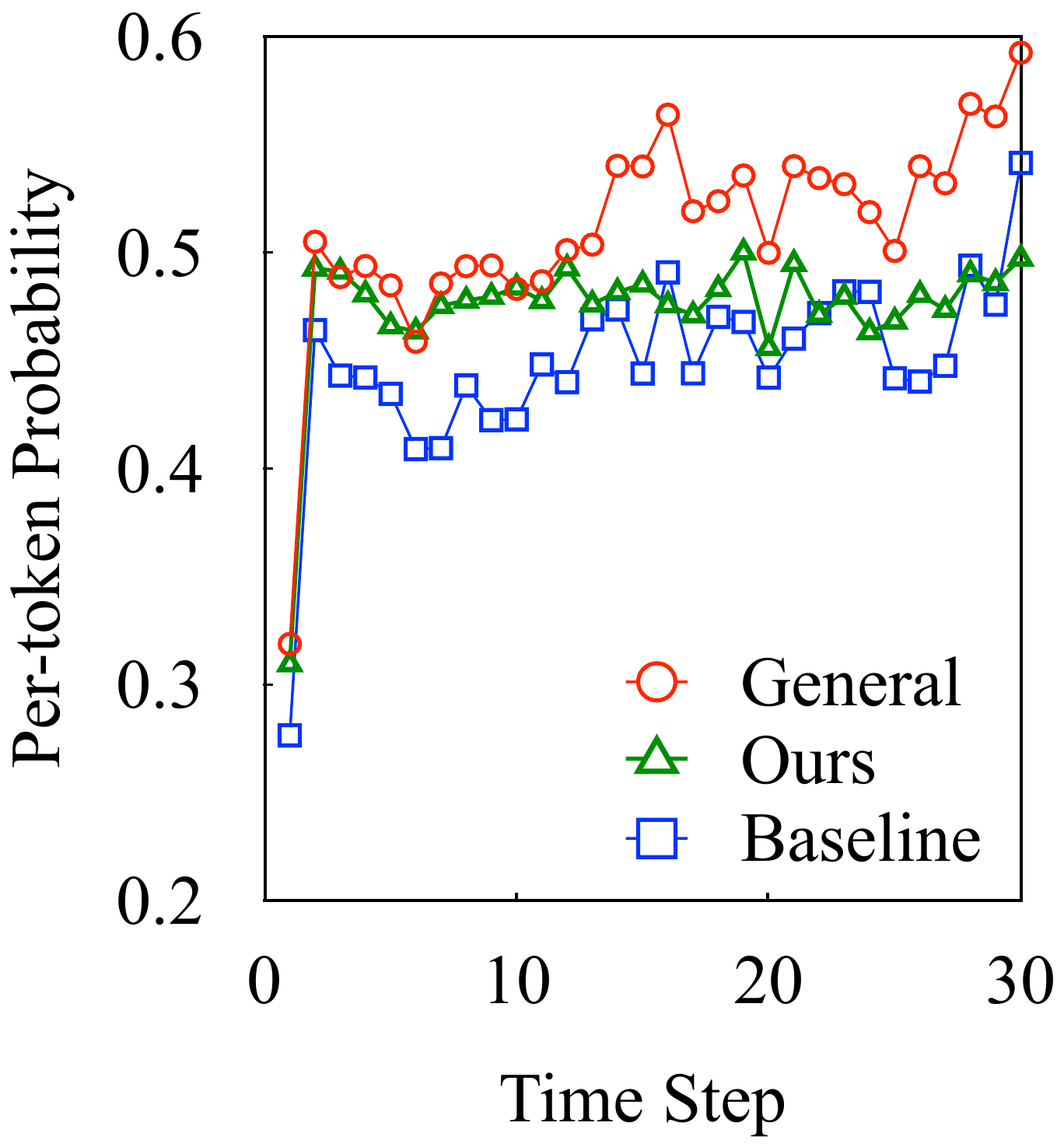}}
    \hfill
    \subfloat[Distractor]{
    \includegraphics[height=0.26\textwidth]{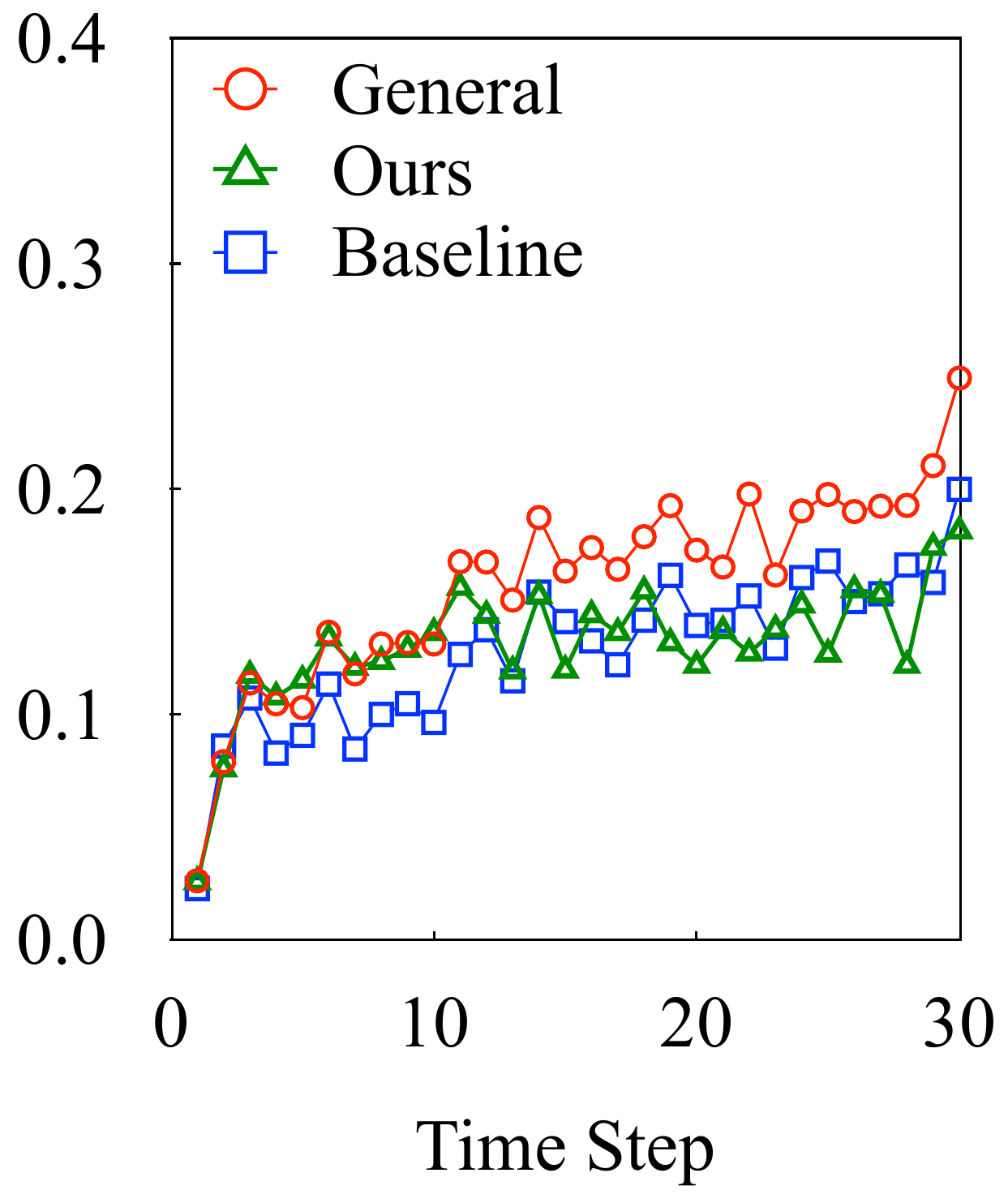}}
    \caption{Per-token generation probability on WMT19 En$\Rightarrow$De~(S) test set when adopting our approaches.}
    \label{fig:model-uncert-impr}
\end{figure}

\paragraph{Alleviating Over-Estimation.}
Figure~\ref{fig:model-uncert-impr} shows the impact of our approach on model uncertainty. 
Clearly, our approach successfully alleviates the over-estimation issue of general pretraining in both the groundtruth and distractor scenarios.

\paragraph{Mitigating Beam Search Degradation.}
We recap the beam search degradation problem with the application of our approaches in Table~\ref{tab:mbart-BSD-impr}. 
The input adaptation approach can noticeably reduce the performance decline when using a larger beam size (e.g., from -1.8 to -0.9), partially due to a reduction of copying tokens in generated translations (e.g., from 19.4\% to 15.3\%). Although in-domain pretraining does not alleviate the beam search degradation problem, it can be combined with input adaptation to build a well-performing NMT system.

\section{Related Work}

\paragraph{Pretraining for NMT.}
Previous pretraining approaches for NMT generally focus on how to effectively integrate pretrained BERT~\cite{Devlin2019BERT} or GPT~\cite{radford2019GPT2} to NMT models. For example,~\newcite{yang2020towards} propose a concerted training framework, and~\newcite{weng2020acquiring} propose a dynamic fusion mechanism and a distillation paradigm to acquire knowledge from BERT and GPT.
In this work, we aim to provide a better understanding of how Seq2Seq pretraining model works for NMT, and propose a simple and effective approach to improve model performance based on these observations.

\paragraph{Intermediate Pretraining.}
Our in-domain pretraining approach is related to recent successes on intermediate pretraining and intermediate task selection in NLU tasks.
For example, \newcite{ye2021influence} investigate the influence of masking policies in intermediate pretraining. \newcite{poth2021pre} explore to select tasks for intermediate pretraining. Closely related to our work, \newcite{gururangan2020dont} propose to continue the pretraining of \textsc{RoBERTa}~\cite{liu2019roberta} on task-specific data.
Inspired by these findings, we employ in-domain pretraining to narrow the domain gap between general Seq2Seq pretraining and NMT training. We also show the necessity of target-side monolingual data on in-domain pretraining (see Appendix~\ref{sec:indomain-langs}), which has not been studied in previous works of in-domain pretraining.

\begin{table}[t]
\centering
\fontsize{10}{12}\selectfont
\begin{tabular}{l cc ll }
\toprule
\multirow{2}{*}{\bf Approach}   
& \multicolumn{2}{c}{\bf BLEU}
& \multicolumn{2}{c}{\bf Copy (\%)}\\
\cmidrule(lr){2-3}  \cmidrule(lr){4-5}
& 5 & 100 & 5  & 100 \\
\midrule
General & 35.3  & \cellcolor{zgreen!90}33.5 & \bf 13.2  & 19.4 \\
~~+ Input Adapt & 35.6  & \cellcolor{zgreen!45}34.7 & 12.5$^\downarrow$  & 15.3$^\downarrow$ \\
~~+ In-Domain & 36.4  & \cellcolor{zgreen!100}33.9 & 12.9  & \bf 19.8 \\
~~~~+ Input Adapt & 36.1 & \cellcolor{zgreen!55}35.0 & 12.6$^\downarrow$ & 15.6$^\downarrow$ \\
\bottomrule
\end{tabular}
\caption{Beam search degradation and ``copy'' translations when adopting our approaches.}
\label{tab:mbart-BSD-impr}
\end{table}

\section{Conclusion}

In this paper we provide a better understanding of Seq2Seq pretraining for NMT by showing both the benefits and side effects. We propose simple and effective approaches to remedy the side effects by bridging the gaps between Seq2Seq pretraining and NMT finetuning, which further improves translation performance and model robustness.
Future directions include validating our findings on more Seq2Seq pretraining models and language pairs.

\bibliography{ref.bib}

\begin{thebibliography}{33}
\expandafter\ifx\csname natexlab\endcsname\relax\def\natexlab#1{#1}\fi

\bibitem[{Belinkov and Bisk(2018)}]{belinkov2018synthetic}
Yonatan Belinkov and Yonatan Bisk. 2018.
\newblock Synthetic and natural noise both break neural machine translation.
\newblock In \emph{ICLR}.

\bibitem[{Cohen and Beck(2019)}]{cohen2019empirical}
Eldan Cohen and Christopher Beck. 2019.
\newblock Empirical analysis of beam search performance degradation in neural
  sequence models.
\newblock In \emph{ICML}.

\bibitem[{Conneau et~al.(2020)Conneau, Khandelwal, Goyal, Chaudhary, Wenzek,
  Guzm{\'a}n, Grave, Ott, Zettlemoyer, and Stoyanov}]{Conneau2020xlmr}
Alexis Conneau, Kartikay Khandelwal, Naman Goyal, Vishrav Chaudhary, Guillaume
  Wenzek, Francisco Guzm{\'a}n, E.~Grave, Myle Ott, Luke Zettlemoyer, and
  Veselin Stoyanov. 2020.
\newblock Unsupervised cross-lingual representation learning at scale.
\newblock In \emph{ACL}.

\bibitem[{Devlin et~al.(2019)Devlin, Chang, Lee, and
  Toutanova}]{Devlin2019BERT}
Jacob Devlin, Ming-Wei Chang, Kenton Lee, and Kristina Toutanova. 2019.
\newblock Bert: Pre-training of deep bidirectional transformers for language
  understanding.
\newblock In \emph{NAACL-HLT}.

\bibitem[{Ding et~al.(2021)Ding, Wang, Liu, Wong, Tao, and
  Tu}]{ding2021understanding}
Liang Ding, Longyue Wang, Xuebo Liu, Derek~F. Wong, Dacheng Tao, and Zhaopeng
  Tu. 2021.
\newblock Understanding and improving lexical choice in non-autoregressive
  translation.
\newblock In \emph{Proc. of ICLR}.

\bibitem[{Du et~al.(2021)Du, Tu, and Jiang}]{Du:2021:ICML}
Cunxiao Du, Zhaopeng Tu, and Jing Jiang. 2021.
\newblock Order-agnostic cross entropy for non-autoregressive machine
  translation.
\newblock In \emph{ICML}.

\bibitem[{Gururangan et~al.(2020)Gururangan, Marasovi{\'c}, Swayamdipta, Lo,
  Beltagy, Downey, and Smith}]{gururangan2020dont}
Suchin Gururangan, Ana Marasovi{\'c}, Swabha Swayamdipta, Kyle Lo, Iz~Beltagy,
  Doug Downey, and Noah~A. Smith. 2020.
\newblock Don{'}t stop pretraining: Adapt language models to domains and tasks.
\newblock In \emph{Proc. of ACL}.

\bibitem[{Jiao et~al.(2020{\natexlab{a}})Jiao, Lyu, and
  King}]{jiao2020exploiting}
Wenxiang Jiao, Michael Lyu, and Irwin King. 2020{\natexlab{a}}.
\newblock Exploiting unsupervised data for emotion recognition in
  conversations.
\newblock In \emph{EMNLP Findings}.

\bibitem[{Jiao et~al.(2020{\natexlab{b}})Jiao, Wang, He, King, Lyu, and
  Tu}]{jiao2020data}
Wenxiang Jiao, Xing Wang, Shilin He, Irwin King, Michael Lyu, and Zhaopeng Tu.
  2020{\natexlab{b}}.
\newblock Data rejuvenation: Exploiting inactive training examples for neural
  machine translation.
\newblock In \emph{EMNLP}.

\bibitem[{Jiao et~al.(2022)Jiao, Wang, He, Tu, King, and Lyu}]{jiao2022data2}
Wenxiang Jiao, Xing Wang, Shilin He, Zhaopeng Tu, Irwin King, and Michael~R
  Lyu. 2022.
\newblock \href {https://doi.org/10.1109/TASLP.2022.3153269} {Exploiting
  inactive examples for natural language generation with data rejuvenation}.
\newblock \emph{IEEE/ACM TASLP}.

\bibitem[{Jiao et~al.(2021)Jiao, Wang, Tu, Shi, Lyu, and
  King}]{Jiao2021SelfTrainingSW}
Wenxiang Jiao, Xing Wang, Zhaopeng Tu, Shuming Shi, Michael~R. Lyu, and Irwin
  King. 2021.
\newblock Self-training sampling with monolingual data uncertainty for neural
  machine translation.
\newblock In \emph{ACL/IJCNLP}.

\bibitem[{Kudo and Richardson(2018)}]{kudo-richardson-2018-sentencepiece}
Taku Kudo and John Richardson. 2018.
\newblock {S}entence{P}iece: A simple and language independent subword
  tokenizer and detokenizer for neural text processing.
\newblock In \emph{Proc. of EMNLP}.

\bibitem[{Lample and Conneau(2019)}]{Lample2019xlm}
Guillaume Lample and Alexis Conneau. 2019.
\newblock Cross-lingual language model pretraining.
\newblock In \emph{NeurIPS}.

\bibitem[{Lee et~al.(2018)Lee, Firat, Agarwal, Fannjiang, and
  Sussillo}]{lee2018hallucinations}
Katherine Lee, Orhan Firat, Ashish Agarwal, Clara Fannjiang, and David
  Sussillo. 2018.
\newblock Hallucinations in neural machine translation.
\newblock In \emph{NeurIPS-IRASL}.

\bibitem[{Lewis et~al.(2020)Lewis, Liu, Goyal, Ghazvininejad, Mohamed, Levy,
  Stoyanov, and Zettlemoyer}]{lewis2020bart}
Mike Lewis, Yinhan Liu, Naman Goyal, Marjan Ghazvininejad, Abdelrahman Mohamed,
  Omer Levy, Veselin Stoyanov, and Luke Zettlemoyer. 2020.
\newblock {BART}: Denoising sequence-to-sequence pre-training for natural
  language generation, translation, and comprehension.
\newblock In \emph{Proc. of ACL}.

\bibitem[{Liu et~al.(2021)Liu, Wang, Wong, Ding, Chao, Shi, and
  Tu}]{Liu:2021:ACL}
Xuebo Liu, Longyue Wang, Derek~F. Wong, Liang Ding, Lidia~S. Chao, Shuming Shi,
  and Zhaopeng Tu. 2021.
\newblock On the copying behaviors of pre-training for neural machine
  translation.
\newblock In \emph{Proc. of ACL}.

\bibitem[{Liu et~al.(2020)Liu, Gu, Goyal, Li, Edunov, Ghazvininejad, Lewis, and
  Zettlemoyer}]{Liu2020mbart}
Yinhan Liu, Jiatao Gu, Naman Goyal, X.~Li, Sergey Edunov, Marjan Ghazvininejad,
  M.~Lewis, and Luke Zettlemoyer. 2020.
\newblock Multilingual denoising pre-training for neural machine translation.
\newblock \emph{TACL}.

\bibitem[{Liu et~al.(2019)Liu, Ott, Goyal, Du, Joshi, Chen, Levy, Lewis,
  Zettlemoyer, and Stoyanov}]{liu2019roberta}
Yinhan Liu, Myle Ott, Naman Goyal, Jingfei Du, Mandar Joshi, Danqi Chen, Omer
  Levy, Mike Lewis, Luke Zettlemoyer, and Veselin Stoyanov. 2019.
\newblock Roberta: A robustly optimized bert pretraining approach.
\newblock \emph{arXiv preprint arXiv:1907.11692}.

\bibitem[{Ott et~al.(2018)Ott, Auli, Grangier, and Ranzato}]{ott2018analyzing}
Myle Ott, Michael Auli, David Grangier, and Marc'Aurelio Ranzato. 2018.
\newblock Analyzing uncertainty in neural machine translation.
\newblock In \emph{Proc. of ICML}.

\bibitem[{Post(2018)}]{post-2018-call}
Matt Post. 2018.
\newblock A call for clarity in reporting {BLEU} scores.
\newblock In \emph{Proceedings of the Third Conference on Machine Translation:
  Research Papers}.

\bibitem[{Poth et~al.(2021)Poth, Pfeiffer, R{\"u}ckl{\'e}, and
  Gurevych}]{poth2021pre}
Clifton Poth, Jonas Pfeiffer, Andreas R{\"u}ckl{\'e}, and Iryna Gurevych. 2021.
\newblock What to pre-train on? efficient intermediate task selection.
\newblock \emph{arXiv}.

\bibitem[{Radford et~al.(2019)Radford, Wu, Child, Luan, Amodei, Sutskever
  et~al.}]{radford2019GPT2}
Alec Radford, Jeffrey Wu, Rewon Child, David Luan, Dario Amodei, Ilya
  Sutskever, et~al. 2019.
\newblock Language models are unsupervised multitask learners.
\newblock \emph{OpenAI blog}.

\bibitem[{Song et~al.(2019)Song, Tan, Qin, Lu, and Liu}]{song2019mass}
Kaitao Song, Xu~Tan, Tao Qin, Jianfeng Lu, and Tie{-}Yan Liu. 2019.
\newblock {MASS:} masked sequence to sequence pre-training for language
  generation.
\newblock In \emph{Proc. of ICML}.

\bibitem[{Tiedemann(2016)}]{tiedemann2016finding}
J{\"o}rg Tiedemann. 2016.
\newblock Finding alternative translations in a large corpus of movie subtitle.
\newblock In \emph{LREC}.

\bibitem[{Tu et~al.(2017{\natexlab{a}})Tu, Liu, Lu, Liu, and Li}]{Tu:2017:TACL}
Zhaopeng Tu, Yang Liu, Zhengdong Lu, Xiaohua Liu, and Hang Li.
  2017{\natexlab{a}}.
\newblock {Context gates for neural machine translation}.
\newblock \emph{TACL}.

\bibitem[{Tu et~al.(2017{\natexlab{b}})Tu, Liu, Shang, Liu, and
  Li}]{Tu:2017:AAAI}
Zhaopeng Tu, Yang Liu, Lifeng Shang, Xiaohua Liu, and Hang Li.
  2017{\natexlab{b}}.
\newblock {Neural machine translation with reconstruction}.
\newblock In \emph{AAAI}.

\bibitem[{Wang and Sennrich(2020)}]{wang2020exposure}
Chaojun Wang and Rico Sennrich. 2020.
\newblock On exposure bias, hallucination and domain shift in neural machine
  translation.
\newblock In \emph{ACL}.

\bibitem[{Wang et~al.(2021)Wang, Tu, Tan, Shi, Sun, and Liu}]{Wang:2021:ACL}
Shuo Wang, Zhaopeng Tu, Zhixing Tan, Shuming Shi, Maosong Sun, and Yang Liu.
  2021.
\newblock On the language coverage bias for neural machine translation.
\newblock In \emph{Proc of ACL}.

\bibitem[{Wang et~al.(2018)Wang, Watanabe, Hughes, Nakagawa, and
  Chelba}]{wang2018denoising}
Wei Wang, Taro Watanabe, Macduff Hughes, Tetsuji Nakagawa, and Ciprian Chelba.
  2018.
\newblock Denoising neural machine translation training with trusted data and
  online data selection.
\newblock In \emph{WMT}.

\bibitem[{Weng et~al.(2020)Weng, Yu, Huang, Cheng, and Luo}]{weng2020acquiring}
Rongxiang Weng, Heng Yu, Shujian Huang, Shanbo Cheng, and Weihua Luo. 2020.
\newblock Acquiring knowledge from pre-trained model to neural machine
  translation.
\newblock In \emph{AAAI}.

\bibitem[{Yang et~al.(2020)Yang, Wang, Zhou, Zhao, Zhang, Yu, and
  Li}]{yang2020towards}
Jiacheng Yang, Mingxuan Wang, Hao Zhou, Chengqi Zhao, Weinan Zhang, Yong Yu,
  and Lei Li. 2020.
\newblock Towards making the most of bert in neural machine translation.
\newblock In \emph{AAAI}.

\bibitem[{Ye et~al.(2021)Ye, Li, Wang, Bolte, Ma, Yih, Ren, and
  Khabsa}]{ye2021influence}
Qinyuan Ye, Belinda~Z Li, Sinong Wang, Benjamin Bolte, Hao Ma, Wen-tau Yih,
  Xiang Ren, and Madian Khabsa. 2021.
\newblock On the influence of masking policies in intermediate pre-training.
\newblock \emph{arXiv}.

\bibitem[{Zhu et~al.(2019)Zhu, Xia, Wu, He, Qin, Zhou, Li, and
  Liu}]{zhu2019incorporating}
Jinhua Zhu, Yingce Xia, Lijun Wu, Di~He, Tao Qin, Wengang Zhou, Houqiang Li,
  and Tieyan Liu. 2019.
\newblock Incorporating bert into neural machine translation.
\newblock In \emph{ICLR}.

\end{thebibliography}
\bibliographystyle{acl_natbib}

\clearpage

\appendix

\section{Appendix}
\label{sec:appendix}

\subsection{Comparison of XLM-R and mBART}
\label{sec:XLMR}

\begin{table}[h]
\centering
\setlength{\tabcolsep}{3.8pt}
\begin{tabular}{cc cc cc cc}
\toprule
\multicolumn{2}{c}{\bf Pre-Train} & \multicolumn{2}{c}{\bf En-De} & \multicolumn{2}{c}{\bf\small En-De (S)} & \multicolumn{2}{c}{\bf En-Ro} \\
\cmidrule(lr){1-2} \cmidrule(lr){3-4} \cmidrule(lr){5-8}
\em {Enc} & \em {Dec} & $\Rightarrow$ & $\Leftarrow$ & $\Rightarrow$ & $\Leftarrow$ & $\Rightarrow$ & $\Leftarrow$ \\
\cmidrule(lr){1-2}  \cmidrule(lr){3-8}
\multicolumn{8}{c}{\bf mBART model}\\
\checkmark & \texttimes & 40.8 & 41.0 & 31.7 & 33.5 & 35.0 & 35.6 \\ 
\checkmark & \checkmark  & 40.8 & \bf 41.4 & \bf 35.3 & \bf 35.7 & \bf 37.1 & \bf 37.4 \\
\midrule
\multicolumn{8}{c}{\bf XLM-R model}\\
\checkmark & \texttimes & \bf 41.6 & \bf 41.4 & 27.7 & 30.1 & 32.8 & 30.0 \\
\checkmark & \checkmark & 41.1 & 40.4 & 31.4 & 32.3  & 34.4 & 33.4  \\ 
\bottomrule
\end{tabular}
\caption{Comparison between 12L-12L mBART and XLM-R in terms of BLEU scores on MT benchmarks.}
\label{tab:benchmark-XLMR}
\end{table}

Throughout our paper, we mainly rely on mBART to investigate the pretrained encoder only setting. Here, we report our results on the same benchmark datasets with the popular pretrained encoder, XLM-R~\cite{Conneau2020xlmr}. We also try to initialize the decoder of NMT models with XLM-R. The results are listed in Table~\ref{tab:benchmark-XLMR}. We find that XLM-R achieves comparable translation performance as mBART on the large-scale WMT19 En-De data but under-performs mBART on small-scale data significantly. The strong results of mBART ensure the reliability of our findings.

\subsection{Domain Classifier}
\label{sec:domain-classifier}
To distinguish general domain and in-domain, we build a domain classifier based on the WMT19 En-De training data and the CC data. We select a subset from the full training data of WMT with some trusted data~\cite{wang2018denoising,jiao2020data} , i.e., WMT newstest2010-2017 consisting of 22404 samples, to reduce the impact of possible noises in the training data. Specifically, we first train a language model on the full WMT training data as the noisy model and then finetune it on the trusted data to obtain the denoised model. For a sentence ${\bf x}$, the difference of confidence between the two models, i.e., $\log P_{noisy}({\bf x}) - \log P_{denoised}({\bf x})$, represents the noise score. We select 1.0M samples with the lowest noise score from the WMT training data and randomly select 1.0M samples from the CC data to train the domain classifier. 
The newstest2018 combined with an equally sized subset of CC data is used as the validation data to select the best classifier.

\subsection{Involved Languages}
\label{sec:indomain-langs}

\begin{table}[h]
  \centering
  \begin{tabular}{l c ccc c}
  \toprule
  \multirow{2}{*}{\bf Lang} &   \multirow{2}{*}{\bf BLEU}    &   \multicolumn{3}{c}{\bf Frequency} \\
  \cmidrule(lr){3-5}  
    &   &   \em Low &  \em Med   &   \em High   &   \\
  \midrule
  None   & 40.8 & 50.9 & 58.0 & 67.0 \\
  En        & 41.3 & 51.1 & 58.1 & 67.5 \\
  En,De     & 42.2 & 52.2 & 59.2 & 67.7 \\
  \bottomrule
  \end{tabular}
  \caption{Effect of languages involved in in-domain pretraining, evaluated on WMT19 En-De dataset.}
  \label{tab:analysis-langs}
\end{table}

We investigate whether the languages involved in the in-domain pretraining process affect the final performance of our approach. In Table~\ref{tab:analysis-langs}, we present the results of in-domain pretraining with only one language involved, i.e., English. While the translation quality can also be improved slightly, the improvements of accuracy on medium- and low-frequency words are very limited. 
It indicates that in-domain pretraining on the target language (i.e., German here) is critical for medium- and low-frequency words.

\end{document}